\documentclass[runningheads]{llncs}
\usepackage{graphicx}
\usepackage{amsmath,amssymb} 
\usepackage{color}
\usepackage[width=122mm,left=12mm,paperwidth=146mm,height=193mm,top=12mm,paperheight=217mm]{geometry}
\usepackage{epsfig}
\usepackage{courier}
\usepackage{multirow}
\usepackage{algorithm}
\usepackage{algorithmic}
\usepackage{breqn}
\usepackage{sidecap}
\usepackage{caption}
\newcommand{\relationship}[3]{$\langle$\texttt{#1} - \texttt{#2} - \texttt{#3}$\rangle$}
\newcommand{\eg}[1]{e.g. #1}
\newcommand\blfootnote[1]{%
  \begingroup
  \renewcommand\thefootnote{}\footnote{#1}%
  \addtocounter{footnote}{-1}%
  \endgroup
}

\begin{document}
\pagestyle{headings}
\mainmatter

\title{Visual Relationship Detection \\ with Language Priors} 

\titlerunning{Visual Relationship Detection with Language Priors}

\authorrunning{Cewu Lu*, Ranjay Krishna*, Michael Bernstein, Li Fei-Fei}

\author{Cewu Lu*, Ranjay Krishna*, Michael Bernstein, Li Fei-Fei \\ 
\small{\{cwlu, ranjaykrishna, msb, feifeili\}@cs.stanford.edu}}

\institute{Stanford University}

\maketitle
\begin{abstract}
Visual relationships capture a wide variety of interactions between pairs of objects in images (\eg ``man riding bicycle'' and ``man pushing bicycle''). Consequently, the set of possible relationships is extremely large and it is difficult to obtain sufficient training examples for all possible relationships. Because of this limitation, previous work on visual relationship detection has concentrated on predicting only a handful of relationships. Though most relationships are infrequent, their objects (\eg ``man'' and ``bicycle'') and predicates (\eg ``riding'' and ``pushing'') independently occur more frequently. We propose a model that uses this insight to train visual models for objects and predicates individually and later combines them together to predict multiple relationships per image. We improve on prior work by leveraging language priors from semantic word embeddings to finetune the likelihood of a predicted relationship. Our model can scale to predict thousands of types of relationships from a few examples. Additionally, we localize the objects in the predicted relationships as bounding boxes in the image. We further demonstrate that understanding relationships can improve content based image retrieval.
\end{abstract}

\section{Introduction}
\begin{figure}[t]
\setlength{\belowcaptionskip}{-20pt}
\centering
\includegraphics[width=0.9\linewidth]{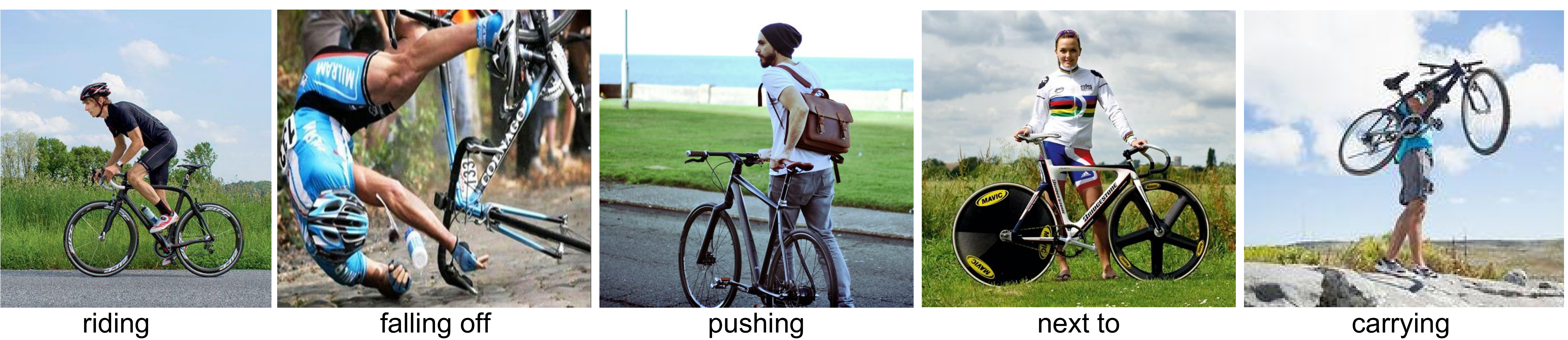}
\caption{Even though all the images contain the same objects (a \texttt{person} and a \texttt{bicycle}), it is the relationship between the objects that determine the holistic interpretation of the image.}
\label{fig:intro_diff}
\end{figure}

\blfootnote{* = equal contribution}
While objects are the core building blocks of an image, it is often the relationships between objects that determine the holistic interpretation. For example, an image with a \texttt{person} and a \texttt{bicycle} might involve the man \texttt{riding}, \texttt{pushing}, or even \texttt{falling off} of the bicycle (Figure~\ref{fig:intro_diff}). Understanding this diversity of relationships is central to accurate image retrieval and to a richer semantic understanding of our visual world.

\begin{figure*}[t]
\centering
\setlength{\belowcaptionskip}{-20pt}
\includegraphics[width=0.9\linewidth]{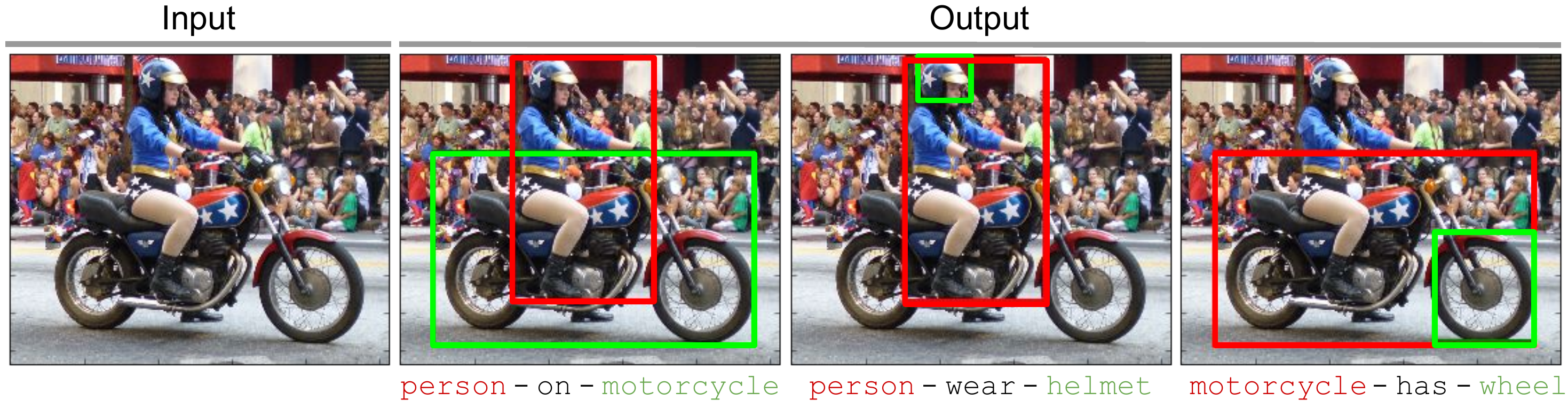}
\caption{Visual Relationship Detection: Given an image as input, we detect multiple relationships in the form of \relationship{object$_1$}{relationship}{object$_2$}. Both the objects are localized in the image as bounding boxes. In this example, we detect the following relationships: \relationship{person}{on}{motorcycle}, \relationship{person}{wear}{helmet} and \relationship{motorcycle}{has}{wheel}.}
\label{fig:task}
\end{figure*}

Visual relationships are a pair of localized objects connected via a predicate (Figure~\ref{fig:task}). We represent relationships as \relationship{object$_1$}{predicate}{object$_2$}~\footnotemark. Visual relationship detection involves detecting and localizing pairs of objects in an image and also classifying the predicate or interaction between each pair (Figure~\ref{fig:task}). While it poses similar challenges as object detection~\cite{everingham2010pascal}, one critical difference is that the size of the semantic space of possible relationships is much larger than that of objects. Since relationships are composed of two objects, there is a greater skew of rare relationships as object co-occurrence is infrequent in images. So, a fundamental challenge in visual relationship detection is learning from very few examples.

\footnotetext{In natural language processing~\cite{zhou122007tree,guodong2005exploring,culotta2004dependency,socher2012semantic}, relationships are defined as \relationship{subject}{predicate}{object}. In this paper, we define them as \relationship{object$_1$}{predicate}{object$_2$} for simplicity.}

Visual Phrases~\cite{sadeghi2011recognition} studied visual relationship detection using a small set of 13 common relationships. Their model requires enough training examples for every possible \relationship{object$_1$}{predicate}{object$_2$} combination, which is difficult to collect owing to the infrequency of relationships. If we have $N$ objects and $K$ predicates, Visual Phrases~\cite{sadeghi2011recognition} would need to train $\mathcal{O}(N^2K)$ unique detectors separately. We use the insight that while relationships (\eg ``person jumping over a fire hydrant'') might occur rarely in images, its objects (\eg \texttt{person} and \texttt{fire hydrant}) and predicate (\eg \texttt{jumping over}) independently appear more frequently. We propose a \textbf{visual appearance module} that learns the appearance of objects and predicates and fuses them together to jointly predict relationships. We show that our model only needs $\mathcal{O}(N + K)$ detectors to detect $\mathcal{O}(N^2K)$ relationships.

Another key observation is that relationships are semantically related to each other. For example, a ``person riding a horse'' and  a ``person riding an elephant'' are semantically similar because both \texttt{elephant} and \texttt{horse} are animals. Even if we haven't seen many examples of ``person riding an elephant'', we might be able to infer it from a ``person riding a horse''. Word vector embeddings~\cite{mikolov2013efficient} naturally lend themselves in linking such relationships because they capture semantic similarity in language (\eg \texttt{elephant} and \texttt{horse} are cast close together in a word vector space). Therefore, we also propose a \textbf{language module} that uses pre-trained word vectors~\cite{mikolov2013efficient} to cast relationships into a vector space where similar relationships are optimized to be close to each other. Using this embedding space, we can finetune the prediction scores of our relationships and even enable zero shot relationship detection.

In this paper, we propose a model that can learn to detect visual relationships by (1) (1) learning visual appearance models for its objects and predicates and  (2) using the relationship embedding space learnt from language. We train our model by optimizing a bi-convex function. To benchmark the task of visual relationship detection, we introduce a new dataset that contains $5000$ images with $37,993$ relationships. Existing datasets that contain relationships were designed for improving object detection~\cite{sadeghi2011recognition} or image retrieval~\cite{Johnson2015Image} and hence, don't contain sufficient variety of relationships or predicate diversity per object category. Our model outperforms all previous models in visual relationship detection. We further study how our model can be used to perform zero shot visual relationship detection. Finally, we demonstrate that understanding relationships can improve image-based retrieval.

\section{Related Work}
Visual relationship prediction involves detecting the objects that occur in an image as well as understanding the interactions between them. There has been a series of work related to improving object detection by leveraging \textbf{object co-occurrence} statistics~\cite{mensink2014costa,salakhutdinov2011learning,ladicky2010graph,rabinovich2007objects,galleguillos2008object,galleguillos2010context}. Structured learning approaches have improved scene classification along with object detection using hierarchial contextual data from co-occurring objects~\cite{choi2010exploiting,izadinia2014incorporating,fidler2007towards,sivic2005discovering}. Unlike these methods, we study the \textit{context} or \textit{relationships} in which these objects co-occur.

Some previous work has attempted to learn \textbf{spatial relationships} between objects~\cite{gould2008multi,galleguillos2008object} to improve segmentation~\cite{gould2008multi}. They attempted to learn four spatial relationships: ``above'', ``below'', ``inside'', and ``around''~\cite{galleguillos2008object}. While we believe that that learning spatial relationships is important, we also study non-spatial relationships such as \texttt{pull} (actions), \texttt{taller than} (comparative), etc.

There have been numerous efforts in \textbf{human-object interaction}~\cite{rohrbach2013translating,yao2010modeling,maji2011action} and action recognition~\cite{gupta2009observing} to learn discriminative models that distinguish between relationships where \texttt{object$_1$} is a human (~\eg ``playing violin''~\cite{yao2010grouplet}). Visual relationship prediction is more general as \texttt{object$_1$} is not constrained to be a human and the \texttt{predicate} doesn't have to be a verb.

\textbf{Visual relationships} are not a new concept. Some papers explicitly collected relationships in images~\cite{ramanathan2015learning,guadarrama2013youtube2text,regneri2013grounding,thomason2014integrating,yao2012describing} and videos~\cite{regneri2013grounding,kulkarni2011baby,zitnick2013learning} and helped models map these relationships from images to language. Relationships have also improved object localization~\cite{gupta2008beyond,kumar2010efficiently,sadeghi2011recognition,russell2006using}. A meaning space of relationships have aided the cognitive task of mapping images to captions~\cite{farhadi2010every,berg2012understanding,hoiem2008putting,fang2014captions}. Finally, they have been used to generate indoor images from sentences~\cite{chang2014semantic} and to improve image search~\cite{Johnson2015Image,schustergenerating}. In this paper, we formalize visual relationship prediction as a task onto itself and demonstrate further improvements in image retrieval.

The most recent attempt at relationship prediction has been in the form of \textbf{visual phrases}. Learning appearance models for visual phrases has shown to improve individual object detection, i.e.\ detecting ``a person riding a horse'' improves the detection and localization of ``person'' and ``horse''~\cite{sadeghi2011recognition,choi2013understanding}.  Unlike our model, all previous work has attempted to detect only a handful of visual relationships and do not scale because most relationships are infrequent. We propose a model that manages to scale and detect millions of types of relationships. Additionally, our model is able to detect unseen relationships.

\begin{figure*}[t]
\centering
\setlength{\belowcaptionskip}{-10pt}
\includegraphics[width=\linewidth]{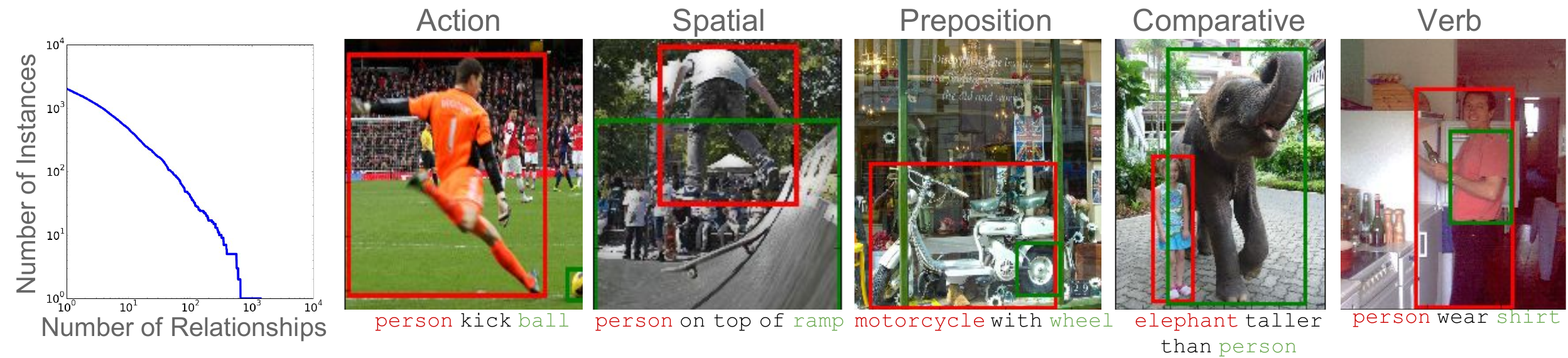}
\vspace{-0.1in}
\caption{(left) A log scale distribution of the number of instances to the number of relationships in our dataset. Only a few relationships occur frequently and there is a long tail of infrequent relationships. (right) Relationships in our dataset can be divided into many categories, 5 of which are shown here: verb, spatial, preposition, comparative and action.}
\label{fig:dataset}
\end{figure*}

\begin{table}[t]
\begin{center}
\small
\caption{Comparison between our visual relationship benchmarking dataset with existing datasets that contain relationships. Relationships and Objects are abbreviated to Rel. and Obj. because of space constraints.}
\label{tab:dataset}
\begin{tabular}{l | c | c | c | c}
  & Images & Rel. Types & Rel. Instances & \# Predicates per Obj. Category\\
  \hline
  Visual Phrases~\cite{sadeghi2011recognition} & 2,769 & 13 & 2,040 & 120\\
  Scene Graph~\cite{Johnson2015Image} & 5,000 & 23,190 & 109,535 & 2.3 \\
  \hline
  Ours & 5,000 & 6,672 & 37,993 & 24.25\\
\end{tabular}
\end{center}
\end{table}

\section{Visual Relationship Dataset}
\label{sec:dataset}
Visual relationships put objects in context; they capture the different interactions between pairs of objects. These interactions (shown in Figure~\ref{fig:dataset}) might be verbs (\eg \texttt{wear}), spatial (\eg \texttt{on top of}), prepositions (\eg \texttt{with}), comparative (\eg \texttt{taller than}), actions (\eg \texttt{kick}) or a preposition phrase (\eg \texttt{drive on}). A dataset for visual relationship prediction is fundamentally different from a dataset for object detection. A relationship dataset should contain more than just objects localized in images; it should capture the rich variety of interactions between pairs of objects (predicates per object category). For example, a \texttt{person} can be associated with predicates such as \texttt{ride}, \texttt{wear}, \texttt{kick} etc. Additionally, the dataset should contain a large number of possible relationships types.

Existing datasets that contain relationships were designed to improve object detection~\cite{sadeghi2011recognition} or image retrieval~\cite{Johnson2015Image}. The Visual Phrases~\cite{sadeghi2011recognition} dataset focuses on 17 common relationship types. But, our goal is to understand the rich variety of infrequent relationships. On the other hand, even though the Scene Graph dataset~\cite{Johnson2015Image} has 23,190 relationship types~\footnotemark, it only has 2.3 predicates per object category. Detecting relationships on the Scene Graph dataset~\cite{Johnson2015Image} essentially boils down to object detection. Therefore, we designed a dataset specifically for benchmarking visual relationship prediction.

Our dataset (Table~\ref{tab:dataset}) contains 5000 images with 100 object categories and 70 predicates. In total, the dataset contains 37,993 relationships with  6,672 relationship types and 24.25 predicates per object category. Some example relationships are shown in Figure~\ref{fig:dataset}. The distribution of relationships in our dataset highlights the long tail of infrequent relationships (Figure~\ref{fig:dataset}(left)). We use 4000 images in our training set and test on the remaining 1000 images. 1,877 relationships occur in the test set but never occur in the training set.

\footnotetext{Note that the Scene Graph dataset~\cite{Johnson2015Image} was collected using unconstrained language, resulting in multiple annotations for the same relationship (\eg \relationship{man}{kick}{ball} and \relationship{person}{is kicking}{soccer ball}). Therefore, 23,190 is an inaccurate estimate of the number of unique relationship types in their dataset. We do not compare with the Visual Genome dataset~\cite{krishnavisualgenome} because their relationships had not been released at the time this paper was written.}

\section{Visual Relationship Prediction Model}
\begin{figure*}[t]
\centering
\setlength{\belowcaptionskip}{-20pt}
\includegraphics[width=0.9\linewidth]{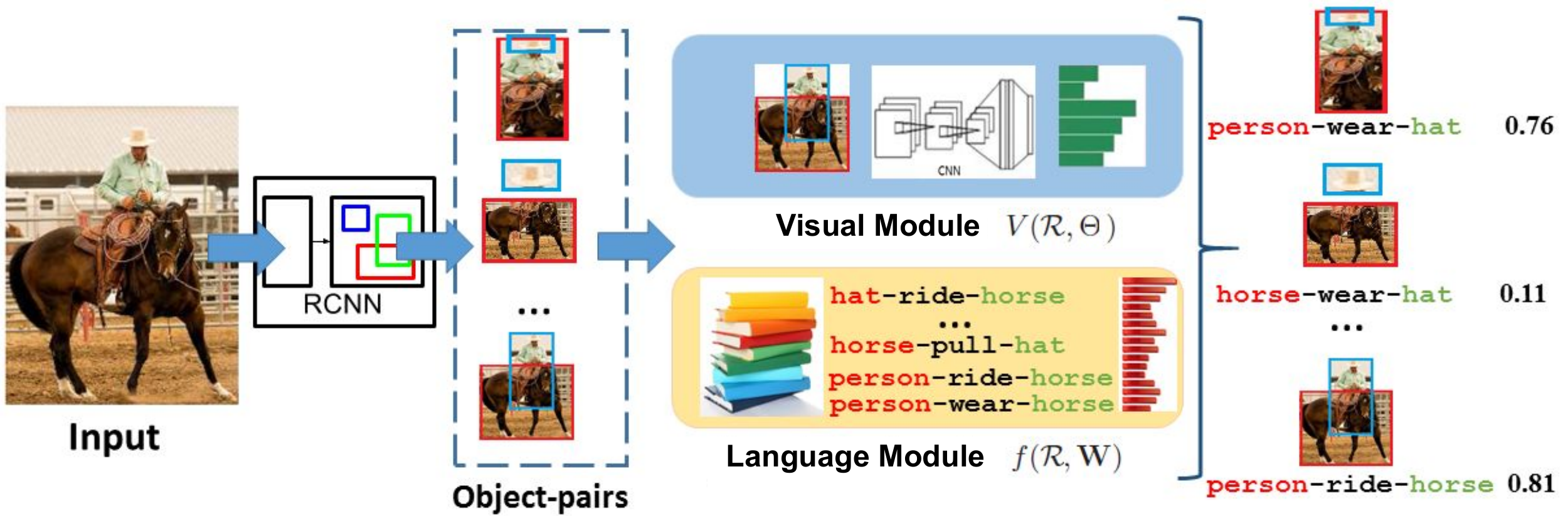}
\caption{A overview of our visual relationship detection pipeline. Given an image as input, RCNN~\cite{girshick14CVPR} generates a set of object proposals. Each pair of object proposals is then scored using a (1) visual appearance module and a (2) language module. These scores are then thresholded to output a set of relationship labels (\eg \relationship{person}{riding}{horse}). Both objects in a relationship (\eg \texttt{person} and \texttt{horse}) are localized as bounding boxes. The parameters of those two modules ($W$ and $\Theta$) are iteratively learnt in Section~\ref{sec:training}.}
\label{fig:system}
\end{figure*}

The goal of our model is to detect visual relationships from an image. During training (Section~\ref{sec:training}), the input to our model is a fully supervised set of images with relationship annotations where the objects are localized as bounding boxes and labelled as \relationship{object$_1$}{predicate}{object$_2$}. At test time (Section~\ref{sec:testing}), our input is an image with no annotations. We predict multiple relationships and localize the objects in the image. Figure~\ref{fig:system} illustrates a high level overview of our detection pipeline.

\subsection{Training Approach}
\label{sec:training}
In this section, we describe how we train our visual appearance and language modules. Both the modules are combined together in our objective function.

\subsubsection{Visual Appearance Module}
\label{sec:scale}
While Visual Phrases~\cite{sadeghi2011recognition} learned a separate detector for every single relationship, we model the appearance of visual relationships $V()$ by learning the individual appearances of its comprising objects and predicate. While relationships are infrequent in real world images, the objects and predicates can be learnt as they independently occur more frequently. Furthermore, we demonstrate that our model outperforms Visual Phrases' detectors, showing that learning individual detectors outperforms learning detectors for relationships together (Table~\ref{tab:relationship_results}).

First, we train a convolutional neural network (CNN) (VGG net~\cite{simonyan2014very}) to classify each of our $N=100$ objects. Similarly, we train a second CNN (VGG net~\cite{simonyan2014very}) to classify each of our $K=70$ predicates using the union of the bounding boxes of the two participating objects in that relationship. Now, for each ground truth relationship $R_{\langle i,k,j \rangle}$ where $i$ and $j$ are the object classes (with bounding boxes $O_1$ and $O_2$) and $k$ is the predicate class, we model $V$ (Figure~\ref{fig:system}) as:
{\small
\begin{eqnarray}
V( R_{\langle i,k,j \rangle}, \Theta | \langle O_1, O_2 \rangle) = P_i(O_1) ( {\bf z}_{k}^{T} \textrm{CNN}(O_1,O_2) + s_k)  P_j(O_2)
\label{eq:visual_model}
\end{eqnarray}
}
where $\Theta$ is the parameter set of $\{ {\bf z}_{k}, s_k\}$. ${\bf z}_{k}$ and $s_k$ are the parameters learnt to convert our CNN features to relationship likelihoods. $k = 1,\ldots, K$ represent the $K$ predicates in our dataset. $P_i(O_1)$ and $P_j(O_2)$ are the CNN likelihoods of categorizing box $O_1$ as object category $i$ and box $O_2$ as category $j$. CNN$(O_1,O_2)$ is the predicate CNN features extracted from the union of the $O_1$ and $O_2$ boxes.

\subsubsection{Language Module}
\label{sec:knowledge}
One of our key observations is that relationships are semantically related to one another. For example, \relationship{person}{ride}{horse} is semantically similar to \relationship{person}{ride}{elephant}. Even if we have not seen any examples of \relationship{person}{ride}{elephant}, we should be able to infer it from similar relationships that occur more frequently (\eg \relationship{person}{ride}{horse}). Our language module projects relationships into an embedding space where similar relationships are optimized to be close together. We first describe the function that projects a relationship to the vector space (Equation~\ref{eq:linkage}) and then explain how we train this function by enforcing similar relationships to be close together in a vector space (Equation~\ref{eq:K}) and by learning a likelihood prior on relationships (Equation~\ref{eq:fre_term}).

\paragraph{Projection Function}
First, we use pre-trained word vectors (word2vec)~\cite{mikolov2013efficient} to cast the two objects in a relationship into an word embedding space~\cite{mikolov2013efficient}. Next, we concatenate these two vectors together and transform it into the relationship vector space using a projection parameterized by ${\bf W}$, which we learn. This projection presents how two objects interact with each other. We denote $\emph{word2vec}()$ as the function that converts a word to its $300$ \textit{dim.} vector. The relationship projection function (shown in Figure~\ref{fig:system}) is defined as:
{\small
\begin{eqnarray}
f(\mathcal{R}_{\langle i,k,j \rangle}, {\bf W})  = {\bf w}_k^{T}[\emph{word2vec}(t_i), \emph{word2vec}(t_j)] + b_k
\label{eq:linkage}
\end{eqnarray}
}
where $t_j$ is the word (in text) of the $j^{th}$ object category. ${\bf w}_k$ is a 600 \textit{dim.} vector and $b_k$ is a bias term.  ${\bf W}$ is the set of $\{ \{ {\bf w}_1, b_1 \}, \ldots, \{ {\bf w}_k, b_k \}\}$, where each row presents one of our K predicates.

\paragraph{Training Projection Function}
We want to optimize the projection function $f()$ such that it projects similar relationships closer to one another. For example, we want the distance between \relationship{man}{riding}{horse} to be close to  \relationship{man}{riding}{cow} but farther from \relationship{car}{has}{wheel}. We formulate this by using a heuristic where the distance between two relationships is proportional to the word2vec distance between its component objects and predicate:
{\small
\begin{eqnarray}
  \frac{[f(\mathcal{R}, {\bf W})  - f(\mathcal{R}^\prime, {\bf W}) ]^2}{d( \mathcal{R}, \mathcal{R}^\prime)}   = constant,  ~~ \forall  \mathcal{R}, \mathcal{R}^\prime
\label{eq:propto}
\end{eqnarray}
}
where $d(\mathcal{R}, \mathcal{R}^\prime)$ is the sum of the cosine distances (in word2vec space~\cite{mikolov2013efficient}) between of the two objects and the predicates of the two relationships $\mathcal{R}$ and $\mathcal{R}^\prime$. Now, to satisfy Eq~\ref{eq:propto}, we randomly sample pairs of relationships ($\langle \mathcal{R},\mathcal{R}^\prime \rangle$) and minimize their variance:
{\small
\begin{eqnarray}
K({\bf W}) = var (\{\frac{[f(\mathcal{R}, {\bf W})  - f(\mathcal{R}^\prime, {\bf W}) ]^2}{d( \mathcal{R}, \mathcal{R}^\prime)}  ~~ \forall \mathcal{R},\mathcal{R}^\prime \})
\label{eq:K}
\end{eqnarray}
}
where $var()$ is a variance function. The sample number we use is $500$K. 

\paragraph{Likelihood of a Relationship}
The output of our projection function should ideally indicate the likelihood of a visual relationship. For example, our model should not assign a high likelihood score to a relationship like \relationship{dog}{drive}{car}, which is unlikely to occur. We model this by enforcing that if $\mathcal{R}$ occurs more frequently than $\mathcal{R}^\prime$ in our training data, then it should have a higher likelihood of occurring again. We formulate this as a rank loss function:

{\small
\begin{eqnarray}
L({\bf W}) = \sum_{\{\mathcal{R},\mathcal{R}^\prime\}} \max \{ f( \mathcal{R}^\prime , {\bf W}) - f(\mathcal{R}, {\bf W}) + 1, 0  \}
\label{eq:fre_term}
\end{eqnarray}
}
While we only enforce this likelihood prior for the relationships that occur in our training data, the projection function $f()$ generalizes it for all \relationship{object$_1$}{predicate}{object$_2$} combinations, even if they are not present in our training data. The $\max$ operator here is to encourage correct ranking (with margin) $f(\mathcal{R}, {\bf W}) - f( \mathcal{R}^\prime , {\bf W}) \geq 1$. Minimizing this objective enforces that a relationship with a lower likelihood of occurring has a lower $f()$ score.

\subsubsection{Objective function}
\label{sec:objective_function}
So far we have presented  our visual appearance module ($V()$) and the language module ($f()$). We combine them to maximize the rank of the ground truth relationship $\mathcal{R}$ with bounding boxes $O_1$ and $O_2$ using the following rank loss function:
{\small
\begin{eqnarray}
C(\Theta, {\bf W}) = \sum_{\langle O_1， O_2 \rangle, \mathcal{R}}  \max \{ 1 - V( \mathcal{R}, \Theta | \langle O_1, O_2 \rangle) f(  \mathcal{R}, {\bf W}) \nonumber\\
+ \max_{ \langle O_1^\prime, O_2^\prime \rangle \ne \langle O_1 , O_2 \rangle, \mathcal{R}^\prime \ne \mathcal{R}} V( \mathcal{R}^\prime, \Theta | \langle O_1^\prime, O_2^\prime \rangle) f(\mathcal{R}^\prime, {\bf W}), 0 \}
\label{eq:visual_rank_loss}
\end{eqnarray}
}

We use a ranking loss function to make it more likely for our model to choose the correct relationship. Given the large number of possible relationships, we find that a classification loss performs worse. Therefore, our final objective function combines Eq~\ref{eq:visual_rank_loss} with Eqs~\ref{eq:K} and~\ref{eq:fre_term} as:
{\small
\begin{eqnarray}
\min_{\Theta, {\bf W}} \{ C(\Theta, {\bf W}) + \lambda_1 L({\bf W}) +  \lambda_2 K({\bf W}) \}
\label{eq:objective}
\end{eqnarray}
}
where $\lambda_1 = 0.05$ and $\lambda_2 = 0.002$ are hyper-parameters that were obtained though grid search to maximize performance on the validation set. Note that both  Eqs~\ref{eq:visual_rank_loss} and~\ref{eq:fre_term} are convex functions. Eq~\ref{eq:K} is a biqudratic function with respect to ${\bf W}$. So our objective function Eq~\ref{eq:objective} has a quadratic closed form. We perform stochastic gradient descent iteratively on Eqs~\ref{eq:visual_rank_loss} and~\ref{eq:fre_term}. It converges in $20 \sim 25$ iterations.

\subsection{Testing} \label{sec:testing}

At test time, we use RCNN~\cite{girshick14CVPR} to produce a set of candidate object proposals for every test image. Next, we use the parameters learnt from the visual appearance model ($\Theta$) and the language module (${\bf W}$) to predict visual relationships ($\mathcal{R}_{\langle i,k,j \rangle}^*$) for every pair of RCNN object proposals $\langle O_1, O_2 \rangle$ using:
\begin{eqnarray}
\mathcal{R}^{*}  = \arg \max_{\mathcal{R}} V(\mathcal{R}, \Theta | \langle O_1, O_2 \rangle)f(\mathcal{R}, {\bf W})
\label{eq:objective_test}
\end{eqnarray}

\section{Experiments}
\label{sec:experiments}
We evaluate our model by detecting visual relationships from images. We show that our proposed method outperforms previous state-of-the-art methods on our dataset (Section~\ref{sec:vre}) as well as on previous datasets (Section~\ref{sec:vpe}). We also measure how our model performs in zero-shot learning of visual relationships (Section~\ref{sec:zsl}). Finally, we demonstrate that understanding visual relationship can improve common computer vision tasks like content based image retrieval (Section~\ref{sec:ibr}).

\begin{figure}[t]
\centering
\setlength{\belowcaptionskip}{-20pt}
\includegraphics[width=0.8\linewidth]{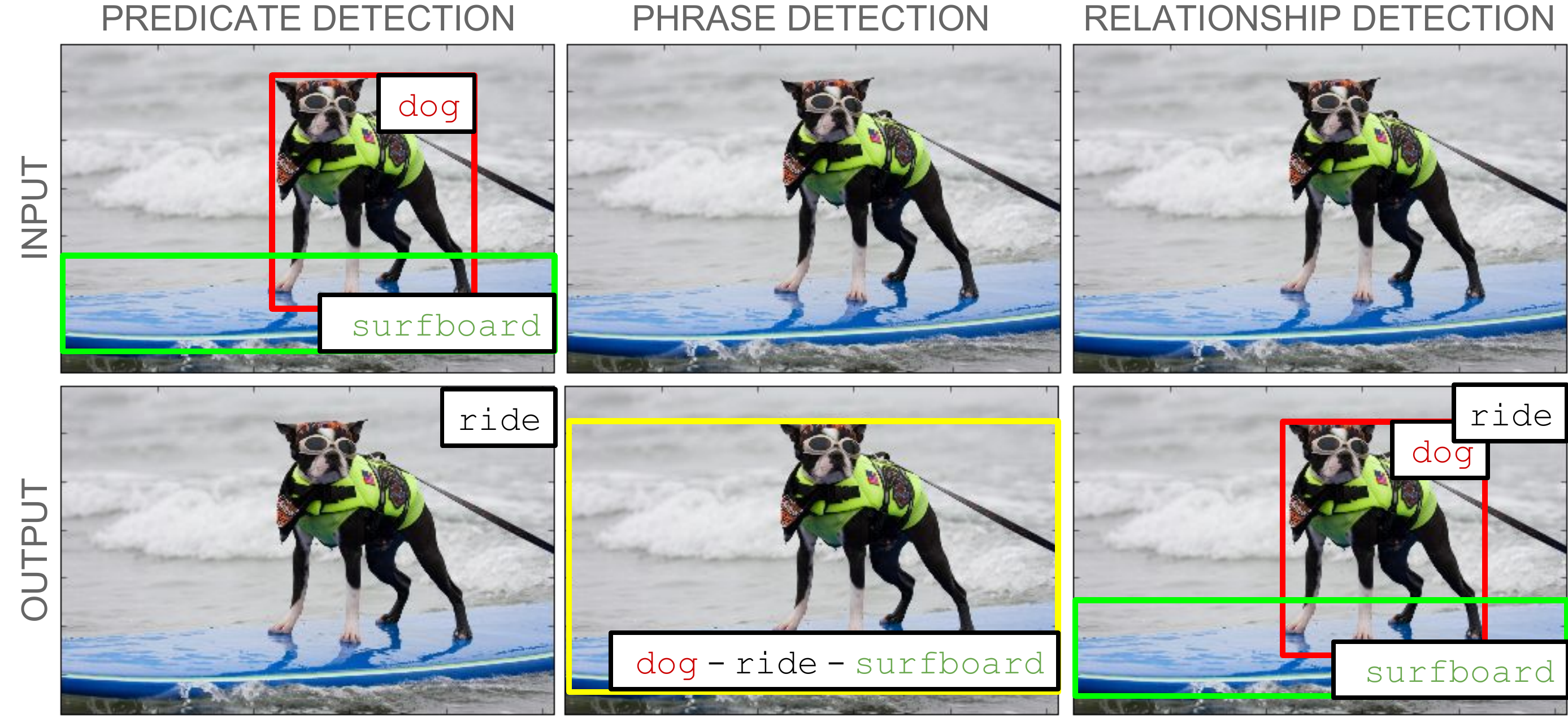}
\caption{We evaluate visual relationship detection using three conditions: predicate detection (where we only predict the predicate given the object classes and boxes), phrase detection (where we label a region of an image with a relationship) and relationship detection (where we detect the objects and label the predicate between them).}
\label{fig:evaluations}
\end{figure}

\subsection{Visual Relationship Detection}
\label{sec:vre}
\paragraph{Setup.} Given an input image, our task is to extract a set of visual relationships  \relationship{object$_1$}{predicate}{object$_2$} and localize the objects as bounding boxes in the image. We train our model using the 4000 training images and perform visual relationship prediction on the 1000 test images.

The evaluation metrics we report is \textbf{recall @ 100} and \textbf{recall @ 50}~\cite{alexe2012measuring}. {\bf Recall @ x} computes the fraction of times the correct relationship is predicted in the top {\bf x} confident relationship predictions. Since we have $70$ predicates and an average of $18$ objects per image, the total possible number of relationship predictions is $100 \times 70 \times 100$, which implies that the random guess will result in a recall @ 100 of $0.00014$. We notice that mean average precision (mAP) is another widely used metric. However, mAP is a pessimistic evaluation metric because we can not exhaustively annotate all possible relationships in an image. Consider the case where our model predicts \relationship{person}{taller than}{person}. Even if the prediction is correct, mAP would penalize the prediction if we do not have that particular ground truth annotation.  

Detecting a visual relationship involves classifying both the objects, predicting the predicate and localization both the objects. To study how our model performs on each of these tasks, we measure visual relationship prediction under the following conditions:

\begin{enumerate}
  \item In \textbf{predicate detection} (Figure~\ref{fig:evaluations}(left)), our input is an image and set of localized objects. The task is to predict a set of possible predicates between pairs of objects. This condition allows us to study how difficult it is to predict relationships without the limitations of object detection~\cite{girshick14CVPR}.
  \item In \textbf{phrase detection} (Figure~\ref{fig:evaluations}(middle)), our input is an image and our task is to output a label \relationship{object$_1$}{predicate}{object$_2$} and localize the entire relationship as \textit{one} bounding box having at least $0.5$ overlap with ground truth box. This is the evaluation used in Visual Phrases~\cite{sadeghi2011recognition}.
  \item In \textbf{relationship detection} (Figure~\ref{fig:evaluations}(right)), our input is an image and our task is to output a set of \relationship{object$_1$}{predicate}{object$_2$} and localize \textit{both} object$_1$ and object$_2$ in the image having at least $0.5$ overlap with their ground truth boxes simultaneously.
\end{enumerate}

\textit{Comparison Models.} We compare our method with some state-of-that-art approaches~\cite{sadeghi2011recognition,simonyan2014very}. We further perform ablation studies on our model, considering just the visual appearance and the language module, including the likelihood term (Eq~\ref{eq:K}) and embedding term (Eq~\ref{eq:fre_term}) to study their contributions.

\begin{itemize}
\setlength\itemsep{0em}
\item \textbf{Visual phrases.} Similar to Visual Phrases~\cite{sadeghi2011recognition}, we train deformable parts models for each of the $6,672$ relationships (\eg ``chair under table'') in our training set.
\item \textbf{Joint CNN.} We train a CNN model~\cite{simonyan2014very} to predict the three components of a relationship together. Specifically, we train a $270$ ($100+100+70$) way classification model that learns to score the two objects (100 categories each) and predicate (70 categories). This model represents the Visual phrases 
\item \textbf{Visual appearance (Ours - V only).} We only use the visual appearance module of our model described in Eq~\ref{eq:visual_rank_loss} by optimizing $V()$.
\item \textbf{Likelihood of a relationship (Ours - L only).} We only use the likelihood of a relationship described in Eq~\ref{eq:fre_term} by optimizing $L()$.
\item \textbf{Visual appearance + naive frequency (Ours - V + naive FC ).} One of the contributions of our model is the ability to use a language prior via our semantic projection function $f()$ (Eq~\ref{eq:linkage}). Here, we replace $f()$ with a function that maps a relationship to its frequency in our training data. Using this naive function, we hope to test the effectiveness of $f()$.
\item \textbf{Visual appearance + Likelihood (Ours - V + L only).} We use both the visual appearance module (Eq~\ref{eq:visual_rank_loss}) and the likelihood term (Eq~\ref{eq:fre_term}) by optimizing both $V()$ and $L()$. The only part of our model missing is $K()$ Eq~\ref{eq:K}, which projects similar relationships closer.
\item \textbf{Visual appearance + likelihood + regularizer (Ours - V + L + Reg.).} We use the visual appearance module and the likelihood term and add an $L_{2}$ regularizer on $W$.
\item \textbf{Full Model (Ours - V + L + K ).} This is our full model. It contains the visual appearance module (Eq~\ref{eq:visual_rank_loss}), the likelihood term (Eq~\ref{eq:fre_term}) and the embedding term (Eq~\ref{eq:K}) from similar relationships.
\end{itemize}

\begin{table}[t]
\centering
\small
\caption{Results for visual relationship detection (Section~\ref{sec:vre}). R@100 and R@50 are abbreviations of Recall @ 100 and Recall @ 50. Note that in predicate det., we are predicting multiple predicates per image (one between every pair of objects) and hence R@100 is less than 1.}
\label{tab:relationship_results}
\begin{tabular}{l  c  c    c  c    c  c   }
         & \multicolumn{2}{c}{Phrase Det.}  &\multicolumn{2}{c}{Relationship Det.} & \multicolumn{2}{c }{Predicate Det.}   \\
         & R@100 &  \multicolumn{1}{c}{R@50} & R@100 &  \multicolumn{1}{c}{R@50} & R@100 &\multicolumn{1}{c }{R@50}  \\
  \hline
  \hline
  Visual Phrases   \cite{sadeghi2011recognition}
                      & 0.07               & 0.04            &  -               & -               &  1.91          &  0.97 \\
  Joint CNN \cite{simonyan2014very}
                      & 0.09               & 0.07            & 0.09             & 0.07            & 2.03           &  1.47  \\
  \hline
  \hline
  Ours - V only       & 2.61               & 2.24            & 1.85             & 1.58            &  7.11          &  7.11  \\
  Ours - L only       & 0.08               & 0.08            &  0.08            & 0.08            &  18.22         &  18.22 \\
  Ours - V + naive FC & 6.39               & 6.65            &  5.47            & 5.27            &  28.87         &  28.87 \\
 Ours - V + L only    & 8.59               &  9.13           &  9.18            & 9.04            &  35.20         &  35.20  \\
 Ours - V + L + Reg.  & 8.91               &  9.60           & 9.63             &  9.71           & 36.31          & 36.31  \\
 Ours - V + L + K     & \textbf{17.03}     & \textbf{16.17}  & \textbf{14.70}   & \textbf{13.86}  & \textbf{47.87} & \textbf{47.87}  \\
\hline
\end{tabular}
\end{table}

\textit{Results.} Visual Phrases~\cite{sadeghi2011recognition} and Joint CNN~\cite{simonyan2014very} train an individual detector for every relationship. Since the space of all possible relationships is large (we have 6,672 relationship types in the training set), there is a shortage of training examples for infrequent relationships, causing both models to perform poorly on predicate, phrase and relationship detection (Table~\ref{tab:relationship_results}). (Ours - V only) can't discriminative between similar relationships by itself resulting in $1.85$ R@100 for relationship detection. Similarly, (Ours - L only) always predicts the most frequent relationship \relationship{person}{wear}{shirt} and results in $0.08$ R@100, which is the percentage of the most frequent relationship in our testing data. These problems are remedied when both V and L are combined in (Ours - V + L only) with an increase of $3\%$ R@100 in on both phrase and relationship detection and more than $10\%$ increase in predicate detection. (V + Naive FC) is missing our relationship projection function $f()$, which learns the likelihood of a predicted relationship and performs worse than (Ours - V + L only) and (Ours - V + L + K). Also, we observe that (Ours - V + L + K) has an $11\%$ improvement in comparison to (Ours - V + L only) in predicate detection, demonstrating that the language module from similar relationships significantly helps improve visual relationship detection. Finally, (Ours - V + L + K) outperforms (Ours - V + L + Reg.) showcasing the $K()$ is acting not only as a regularizer but is learning to preserve the distances between similar relationships.

\begin{figure}[t]
\centering
\setlength{\belowcaptionskip}{-20pt}
\includegraphics[width=0.75\linewidth]{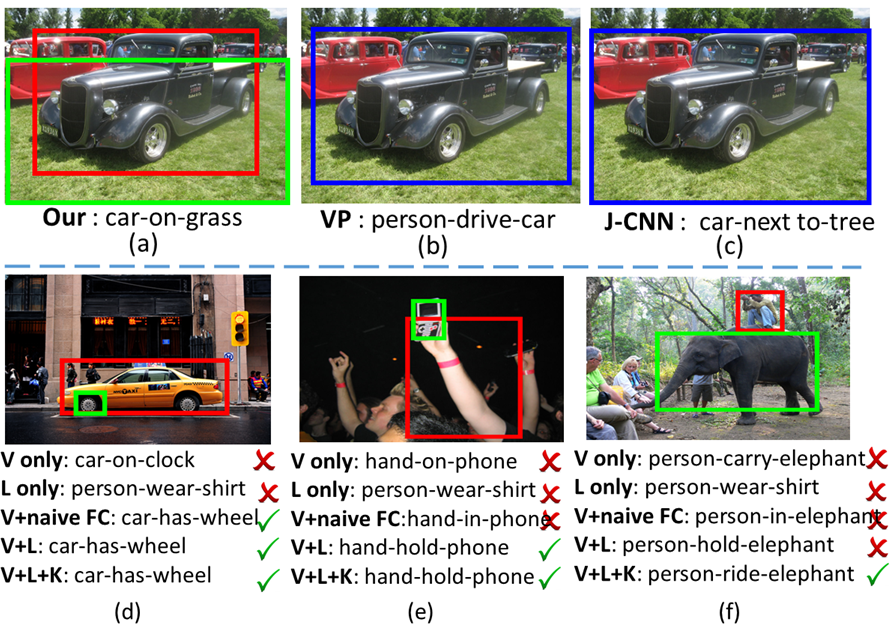}
\vspace{-0.1in}
\caption{(a), (b) and (c) show results from our model, Visual Phrases~\cite{sadeghi2011recognition} and Joint CNN \cite{simonyan2014very} on the same image. All ablation studies results for (d), (e) and (f) are reported below the corresponding image. Ticks and crosses mark the correct and incorrect results respectively. Phrase, object$_1$ and object$_2$ boxes are in blue, red and green respectively.}
\label{fig:relationship_results}
\end{figure}

By comparing the performance of all the models between relationship and predicate detection, we notice a $30\%$ drop in R@100. This drop in recall is largely because we have to localize two objects simultaneously, amplifying the object detection errors. Note that even when we have ground truth object proposals (in predicate detection), R@100 is still $47.87$. 

\textit{Qualitative Results.} In Figure~\ref{fig:relationship_results}(a)(b)(c), Visual Phrase and Joint CNN incorrectly predict a common relationship: \relationship{person}{drive}{car} and \relationship{car}{next to}{tree}. These models tend to predict the most common relationship as they see a lot of them during training. In comparison, our model correctly predicts and localizes the objects in the image. Figure~\ref{fig:relationship_results}(d)(e)(f) compares the various components of our model. Without the relationship likelihood score, (Ours - V only) incorrectly classifies a \texttt{wheel} as a clock in (d) and mislabels the predicate in (e) and (f). Without any visual priors, (Ours - L only) always reports the most frequent relationship \relationship{person}{wear}{shirt}. (Ours - V + L) fixes (d) by correcting the visual model's misclassification of the \texttt{wheel} as a \texttt{clock}. But it still does not predict the correct predicate for (e) and (f) because \relationship{person}{ride}{elephant} and \relationship{hand}{hold}{phone} rarely occur in our training set. However, our full model (Ours - V + L + K) leverages similar relationships it has seen before 
and is able to correctly detect the relationships in (e) and (f).

\begin{table}[t]
\centering
\small
\setlength{\tabcolsep}{2pt}
\caption{Results for zero-shot visual relationship detection (Section~\ref{sec:zero_shot}). Visual Phrases, Joint CNN and Ours - V + naive FC are omitted from this experiment as they are unable to do zero-shot learning.}
\begin{tabular}{l  c  c  c  c  c  c c c c }
         & \multicolumn{2}{c}{Phrase Det.}  &\multicolumn{2}{c}{Relationship Det.} & \multicolumn{2}{c}{Predicate Det.}   \\
         & R@100 &   \multicolumn{1}{c}{R@50} & R@100   &\multicolumn{1}{c}{R@50} & R@100   & \multicolumn{1}{c}{R@50}  \\
        \hline
        \hline
        Ours - V only      & 1.12          & 0.95          &   0.78        & 0.67          & 3.52          &  3.52 \\
        Ours - L only      & 0.01          & 0.00          &   0.01        & 0.00          & 5.09          &  5.09  \\
        Ours - V + L only  & 2.56          & 2.43          &  2.66         & 2.27          & 6.11          &  6.11  \\
        Ours - V + L + K   &\textbf{3.75}  &\textbf{3.36}  &\textbf{3.52}  &\textbf{3.13}  &\textbf{8.45}  &\textbf{8.45}  \\
        \hline
\end{tabular}
\label{tab:relationship_results_zero_shot}
\end{table}

\subsection{Zero-shot Learning}
\label{sec:zsl}
Owing to the long tail of relationships in real world images, it is difficult to build a dataset with every possible relationship. Therefore, a model that detects visual relationships should also be able to perform zero-shot prediction of relationships it has never seen before. Our model is able to leverage similar relationships it has already seen to detect unseen ones.

\label{sec:zero_shot}
\textit{Setup.} Our test set contains $1,877$ relationships that never occur in our training set (\eg \relationship{elephant}{stand on}{street}). These unseen relationships can be inferred by our model using similar relationships (\eg \relationship{dog}{stand on}{street}) from our training set. We report our results for detecting unseen relationships in Table~\ref{tab:relationship_results_zero_shot} for predicate, phrase, and relationship detection. 

\textit{Results.} 
(Ours - V) achieves a low $3.52$ R@100 in predicate detection because visual appearances are not discriminative enough to predict unseen relationships. (Ours - L only)  performs poorly in predicate detection ($5.09$ R@100) because it automatically returns the most common predicate. 
By comparing (Ours - V + L+ K) and (Ours - V + L only), we find the use of K gains an improvement of $30\%$ since it utilizes similar relationships to enable zero shot predictions.

\subsection{Visual Relationship Detection on Existing Dataset}
\label{sec:vpe}
Our goal in this paper is to understand the rich variety of infrequent relationships. Our comparisons in Section~\ref{sec:dataset} show that existing datasets either do not have enough diveristy of predicates per object category or enough relationship types. Therefore, we introduced a new dataset (in Section~\ref{sec:dataset}) and tested our visual relationship detection model in Section~\ref{sec:vre} and Section~\ref{sec:zsl}. In this section, we run additional experiments on the existing visual phrases dataset~\cite{sadeghi2011recognition} to provide further benchmarks.

\begin{table}[t]
\centering
\small
\caption{Visual phrase detection results on Visual Phrases dataset~\cite{sadeghi2011recognition}.}
\begin{tabular}{l c c c c c c}
                      & \multicolumn{3}{c}{Phrase Detection}  &  \multicolumn{3}{c}{Zero-Shot Phrase Detection}  \\
                     & R@100       & R@50        & mAP         & R@100       & R@50         & mAP \\
  \hline
  \hline
  Visual Phrase \cite{sadeghi2011recognition}
                     & 52.7        & 49.3        & 38.0        & -           & -            & -   \\
  Joint CNN          & 75.3        & 71.5        & 54.1        & -           & -            & -   \\
  \hline
  \hline
Ours V only          & 72.0        & 68.6        & 53.4        & 13.5        & 11.3         & 5.3 \\
Ours  V + naive FC   & 77.8        & 73.4        & 55.8        & -            & -           & -   \\
Ours  V + L only     & 79.3        & 76.7        & 57.3        & 17.8        & 15.1         & 8.8 \\
Ours V + L + K       &\textbf{82.7}&\textbf{78.1}&\textbf{59.2}&\textbf{11.4}&\textbf{23.9}&\textbf{18.5}\\
\hline
\end{tabular}
\label{tab:phrase_results}
\end{table}

\textit{Setup.} The visual phrase dataset contains 17 phrases (\eg ``dog jumping''). We evaluate the models (introduced in Section~\ref{sec:vre}) for visual relationship detection on 12 of these phrases that can be represented as a \relationship{object$_1$}{predicate}{object$_2$} relationship. To study zero-shot learning, we remove two phrases (``person lying on sofa'' and ``person lying on beach'') from the training set, and attempt to recognize them in the testing set. We report mAP, R@50 and R@100.

\textit{Results.} In Table~\ref{tab:phrase_results} we see that our method is able to perform better than the existing Visual Phrases' model even though the dataset is small  and contains only $12$ relationships. We get a mAP of $0.59$ using our entire model as compared to a mAP of $0.38$ using Visual Phrases' model. We also outperform the Joint CNN baseline, which achieves a mAP of $0.54$. Considering that (Ours - V only) model performs similarly to the baselines, we believe that our full model's improvements on this dataset are heavily influenced by the language priors.  By learning to embed similar relationships close to each other, the language model's aid can be thought of as being synonymous to the improvements achieved through training set augmentation. Finally, we see a similar improvements in zero shot learning. 

\subsection{Image based Retrieval}
\label{sec:ibr}

An important task in computer vision is image retrieval. An improved retrieval model should be able to infer the relationships between objects in images. We will demonstrate that the use of visual relationships can improve retrieval quality.

\textit{Setup.} Recall that our test set contains $1000$ images. Every query uses $1$ of these $1000$ images and ranks the remaining $999$. We use $54$ query images in our experiments. Two annotators were asked to rank image results for each of the $54$ queries.  To avoid bias, we consider the results for a particular query as ground truth only if it was selected by both annotators. We evaluate performance using R@1, R@5 and R@10 and median rank~\cite{Johnson2015Image}. For comparison, we use three image descriptors that are commonly used in image retrieval: CNN~\cite{simonyan2014very}, GIST~\cite{oliva2001modeling} and SIFT~\cite{lowe2004distinctive}. We rank results for a query using the $L_2$ distance from the query image.
Given a query image, our model predicts a set of visual relationships $\{ R_{1}, \ldots, R_{n} \}$ with a probability of $\{ P_{1}^{q}, \ldots, P_{n}^{q}\}$ respectively. Next, for every image $I_i$ in our test set, it predicts ${R_{1}, \ldots, R_{n}}$ with a confidence of $\{ P_{1}^{i}, \ldots, P_{n}^{i}\}$. We calculate a matching score between an image with the query as $\sum_{j =1}^{n} P_{j}^{q} * P_{j}^{i}$. We also compare our model with Visual Phrases' detectors~\cite{sadeghi2011recognition}.

\begin{figure*}[t]
\centering
\setlength{\belowcaptionskip}{-15pt}
\includegraphics[width=0.85\linewidth]{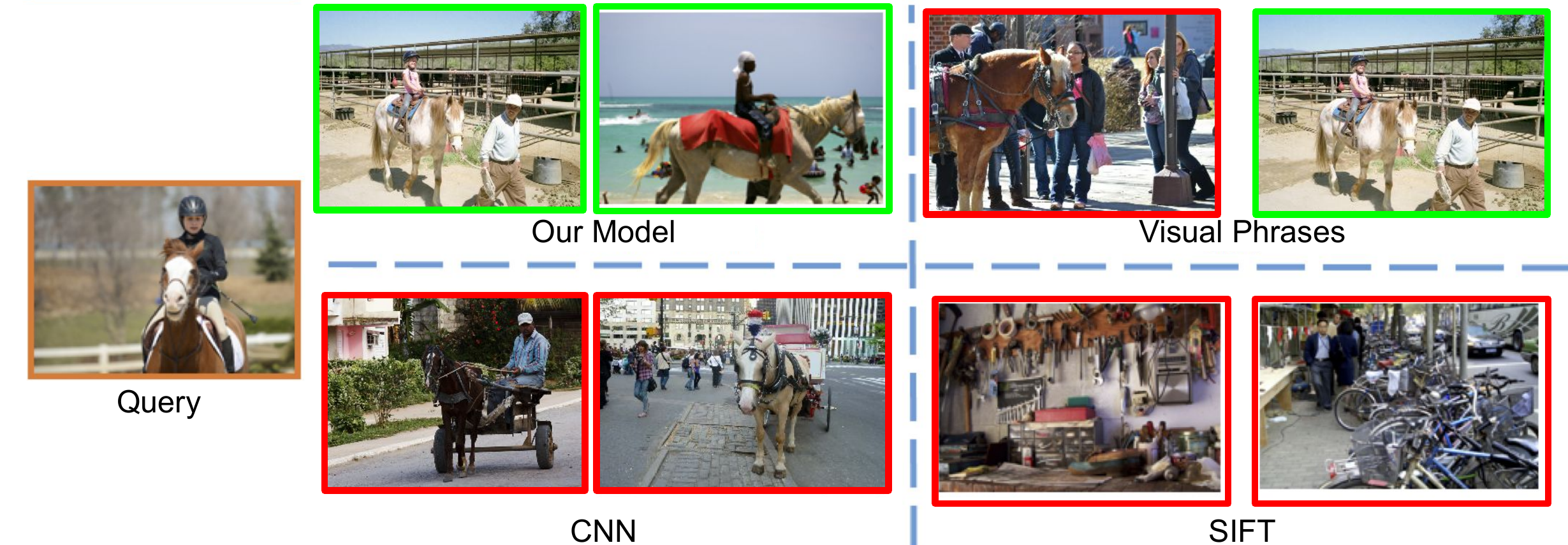}
\caption{Examples retrieval results using an image as the query. }
\label{fig:retrieval_results}
\end{figure*}

\begin{table}[t]
\centering
\small
\caption{Example image retrieval using a image of a \relationship{person}{ride}{horse} (Section~\ref{sec:ibr}). Note that a \textit{higher} recall and \textit{lower} median rank indicates better performance.}
\begin{tabular}{l c c c c}
     & Recall @ 1 & Recall @ 5 & Recall @ 10 & Median Rank\\
  \hline
  \hline
  GIST~\cite{oliva2001modeling} & 0.00 & 5.60 & 8.70 & 68\\
  SIFT~\cite{lowe2004distinctive} & 0.70 & 6.10 & 10.3 & 54\\
  CNN~\cite{simonyan2014very} & 3.15 & 7.70 & 11.5 & 20\\
  Visual Phrases~\cite{sadeghi2011recognition} & 8.72 & 18.12 & 28.04 & 12 \\
  \hline
  \hline
  Our Model & \textbf{10.82} & \textbf{30.02} & \textbf{47.00} & \textbf{4}\\
\end{tabular}
\label{tab:image_retrieval}
\end{table}

\textit{Results.} SIFT~\cite{lowe2004distinctive} and GIST~\cite{oliva2001modeling} descriptors perform poorly with a median rank of $54$ and $68$ (Table~\ref{tab:image_retrieval}) because they simply measure structural similarity between images.
CNN~\cite{simonyan2014very} descriptors capture object-level information and performs better with a median rank of $20$.
Our method captures the visual relationships present in the query image, which is important for high quality image retrieval, improving with a median rank of $4$. When queried using an image of a ''person riding a horse'' (Figure~\ref{fig:retrieval_results}), SIFT returns images that are visually similar but are not semantically relevant. CNN retrieves one image that contains a horse and one that contains both a man and a horse but neither of them capture the relationship: ``person riding a horse''. Visual Phrases and our model are able to detect the relationship \relationship{person}{ride}{horse} and perform better.

\section{Conclusion}
We proposed a model to detect multiple visual relationships in a single image. Our model learned to detect thousands of relationships even when there were very few training examples. We learned the visual appearance of objects and predicates and combined them to predict relationships. To finetune our predictions, we utilized a language prior that mapped similar relationships together -- outperforming previous state of the art~\cite{sadeghi2011recognition} on the visual phrases dataset~\cite{sadeghi2011recognition} as well as our dataset. We also demonstrated that our model can be used for zero shot learning of visual relationships. We introduced a new dataset with $37,993$ relationships that can be used for further benchmarking. Finally, by understanding visual relationships, our model improved content based image retrieval.

\section{Supplementary Material}

\subsection{Training Algorithm}
While we describe the theory and training procedure in the main text of this paper (Section 4.1), we include an algorithm box (Algorithm~\ref{alg:0}) to explain our training procedure in an alternate format.

\begin{algorithm}[bpt]
\caption{Training Algorithm}
\begin{algorithmic}[1]
\STATE {\bf input}: training set of images with annotated \relationship{subject}{predicate}{object} relationships annotated
\STATE Train object detectors on images using RCNN~\cite{girshick14CVPR}
\STATE Train predicate classifier on images using VGG~\cite{simonyan2014very}
\STATE Initialize $f({\bf W})$ (Eq.~2) with word vectors for objects using $\emph{word2vec}()$~\cite{mikolov2013efficient}
\REPEAT 
    \STATE Compute the visual appearance model $V(\Theta)$ (Eq.~1)
    \STATE Compute relationship semantic distance to build $K({\bf W})$ (Eq.~4)
    \STATE Compute the likelihood score $L({\bf W})$ (Eq.~5)
    \STATE Backpropagate and optimize $\{\Theta, {\bf W}\}$ (Eq.~7) using stochastic gradient descent
\UNTIL{$\{\Theta, {\bf W}\}$ have converged}
\STATE {\bf output}: $\{\Theta, {\bf W}\}$
\end{algorithmic}\label{alg:0}
\end{algorithm}

\subsection{Mean Average Precision on Visual Relationship Detection}
As discussed in our paper, mean average precision (mAP) is a pessimistic evaluation metric for visual relationship detection because our dataset does not exhaustively annotate every possible relationship between two pairs of objects. For example, consider the case when our model predicts that a \relationship{person}{next to}{bicycle} when the ground truth annotation is \relationship{person}{push}{bicycle}. In such a case, the prediction is not incorrect but would be penalized by mAP. However, to facilitate future comparisons against our model using this dataset, we report the mAP scores in Table \ref{tab:map_relationship_results}.

We see a similar trend in the mAP scores as we did with the recall @ 50 and recall @ 100 values. The Visual Phrases~\cite{sadeghi2011recognition} and Joint CNN baselines along with (Ours - L only) perform poorly on all three tasks: phrase, relationship and predicate detection. The visual only model (Ours - V only) improved upon these results by leveraging the visual appearances of objects to aid it's predicate detection. Our complete model (Ours - V + L + K) achieves a mAP of $1.52$ on relationship predication since it is penalized for missing annotations. However, it still performs better than all the other ablated models. It also attains a $29.47$ mAP on predicate detection, demonstrating that our model learns to recognize predicates from one another.

\begin{table}[t]
\centering
\small
\caption{mAP results for visual relationship detection (Section~$5.1$).}
\label{tab:map_relationship_results}
\begin{tabular}{l  c  c    c}
         & Phrase Detection  & Relationship Detection & Predicate Detection \\
  \hline
  \hline
  Visual Phrases~\cite{sadeghi2011recognition}
                      & 0.03               & -               &   0.71 \\
  Joint CNN~\cite{simonyan2014very}
                      & 0.05               & 0.04            & 1.02     \\
  \hline
  \hline
  Ours - V only       & 0.93               & 0.84            & 6.42      \\
  Ours - L only       & 0.08               & 0.08            & 8.94      \\
  Ours - V + naive FC & 1.21               & 1.19            & 11.05    \\
 Ours - V + L only    & 1.74               &  1.32           & 16.31   \\
 Ours - V + L + Reg.  & 1.78               &  1.40           & 17.95   \\
 Ours - V + L + K     & \textbf{2.07}      & \textbf{1.52}   & \textbf{29.47}   \\
\hline
\end{tabular}
\end{table}

\subsection{Mean Average Precision on Zero-shot Learning}
Similar to the previous section, we also include the mAP scores for zero shot learning in Table~\ref{tab:map_zero_results}. Again, we see that the the inclusion of K() allows our model to levearage similar relationships to improve zero shot learning in all three experiments.

\begin{table}[t]
\begin{center}
\setlength{\tabcolsep}{4pt}
\caption{mAP results for zero-shot visual relationship detection (Section~$5.2$).}
\label{tab:map_zero_results}
\begin{tabular}{l c c c }
         & Phrase Detection & Relationship Detection & Predicate Detection \\
        \hline
        \hline
        Ours - V only  & 0.92 & 1.03 & 2.13\\
        Ours - L only  & 0.00 & 0.00 & 3.31\\
        Ours - V + L only & 1.97 & 2.30 & 4.45\\
        Ours - V + L + K & \textbf{2.89} &  \textbf{3.01} & \textbf{5.52}\\
        \hline
\end{tabular}
\end{center}
\end{table}

\subsection{Human Evaluation on our Dataset}
We ran an experiment to evaluate the human performance on our dataset. We randomly selecting 1000 pairs of objects from the dataset and then asked humans on Amazon Mechanical Turk to decide which of the $70$ predicates were correct for each pair. We found that humans managed a $98.1\%$ recall @ 50 and $96.4\%$ mAP. This demonstrates that while this task is easy for humans, Visual Relationship Detection is still a hard unsolved task.

\paragraph{Acknowledgements}. Our work is partially funded by an ONR MURI grant.

\bibliographystyle{splncs}
\bibliography{egbib}

\begin{thebibliography}{10}

\bibitem{everingham2010pascal}
Everingham, M., Van~Gool, L., Williams, C.K., Winn, J., Zisserman, A.:
\newblock The pascal visual object classes (voc) challenge.
\newblock International journal of computer vision \textbf{88}(2) (2010)
  303--338

\bibitem{zhou122007tree}
ZHOU12, G., Zhang, M., Ji, D.H., Zhu, Q.:
\newblock Tree kernel-based relation extraction with context-sensitive
  structured parse tree information.
\newblock EMNLP-CoNLL 2007 (2007)  728

\bibitem{guodong2005exploring}
GuoDong, Z., Jian, S., Jie, Z., Min, Z.:
\newblock Exploring various knowledge in relation extraction.
\newblock In: Proceedings of the 43rd annual meeting on association for
  computational linguistics, Association for Computational Linguistics (2005)
  427--434

\bibitem{culotta2004dependency}
Culotta, A., Sorensen, J.:
\newblock Dependency tree kernels for relation extraction.
\newblock In: Proceedings of the 42nd Annual Meeting on Association for
  Computational Linguistics, Association for Computational Linguistics (2004)
  423

\bibitem{socher2012semantic}
Socher, R., Huval, B., Manning, C.D., Ng, A.Y.:
\newblock Semantic compositionality through recursive matrix-vector spaces.
\newblock In: Proceedings of the 2012 Joint Conference on Empirical Methods in
  Natural Language Processing and Computational Natural Language Learning,
  Association for Computational Linguistics (2012)  1201--1211

\bibitem{sadeghi2011recognition}
Sadeghi, M.A., Farhadi, A.:
\newblock Recognition using visual phrases.
\newblock In: Computer Vision and Pattern Recognition (CVPR), 2011 IEEE
  Conference on, IEEE (2011)  1745--1752

\bibitem{mikolov2013efficient}
Mikolov, T., Chen, K., Corrado, G., Dean, J.:
\newblock Efficient estimation of word representations in vector space.
\newblock arXiv preprint arXiv:1301.3781 (2013)

\bibitem{Johnson2015Image}
Johnson, J., Krishna, R., Stark, M., Li, L.J., Shamma, D.A., Bernstein, M.,
  Fei-Fei, L.:
\newblock Image retrieval using scene graphs.
\newblock In: IEEE Conference on Computer Vision and Pattern Recognition
  (CVPR). (2015)

\bibitem{mensink2014costa}
Mensink, T., Gavves, E., Snoek, C.G.:
\newblock Costa: Co-occurrence statistics for zero-shot classification.
\newblock In: Computer Vision and Pattern Recognition (CVPR), 2014 IEEE
  Conference on, IEEE (2014)  2441--2448

\bibitem{salakhutdinov2011learning}
Salakhutdinov, R., Torralba, A., Tenenbaum, J.:
\newblock Learning to share visual appearance for multiclass object detection.
\newblock In: Computer Vision and Pattern Recognition (CVPR), 2011 IEEE
  Conference on, IEEE (2011)  1481--1488

\bibitem{ladicky2010graph}
Ladicky, L., Russell, C., Kohli, P., Torr, P.H.:
\newblock Graph cut based inference with co-occurrence statistics.
\newblock In: Computer Vision--ECCV 2010.
\newblock Springer (2010)  239--253

\bibitem{rabinovich2007objects}
Rabinovich, A., Vedaldi, A., Galleguillos, C., Wiewiora, E., Belongie, S.:
\newblock Objects in context.
\newblock In: Computer vision, 2007. ICCV 2007. IEEE 11th international
  conference on, IEEE (2007)  1--8

\bibitem{galleguillos2008object}
Galleguillos, C., Rabinovich, A., Belongie, S.:
\newblock Object categorization using co-occurrence, location and appearance.
\newblock In: Computer Vision and Pattern Recognition, 2008. CVPR 2008. IEEE
  Conference on, IEEE (2008)  1--8

\bibitem{galleguillos2010context}
Galleguillos, C., Belongie, S.:
\newblock Context based object categorization: A critical survey.
\newblock Computer Vision and Image Understanding \textbf{114}(6) (2010)
  712--722

\bibitem{choi2010exploiting}
Choi, M.J., Lim, J.J., Torralba, A., Willsky, A.S.:
\newblock Exploiting hierarchical context on a large database of object
  categories.
\newblock In: Computer vision and pattern recognition (CVPR), 2010 IEEE
  conference on, IEEE (2010)  129--136

\bibitem{izadinia2014incorporating}
Izadinia, H., Sadeghi, F., Farhadi, A.:
\newblock Incorporating scene context and object layout into appearance
  modeling.
\newblock In: Computer Vision and Pattern Recognition (CVPR), 2014 IEEE
  Conference on, IEEE (2014)  232--239

\bibitem{fidler2007towards}
Fidler, S., Leonardis, A.:
\newblock Towards scalable representations of object categories: Learning a
  hierarchy of parts.
\newblock In: Computer Vision and Pattern Recognition, 2007. CVPR'07. IEEE
  Conference on, IEEE (2007)  1--8

\bibitem{sivic2005discovering}
Sivic, J., Russell, B.C., Efros, A., Zisserman, A., Freeman, W.T.,  et~al.:
\newblock Discovering objects and their location in images.
\newblock In: Computer Vision, 2005. ICCV 2005. Tenth IEEE International
  Conference on. Volume~1., IEEE (2005)  370--377

\bibitem{gould2008multi}
Gould, S., Rodgers, J., Cohen, D., Elidan, G., Koller, D.:
\newblock Multi-class segmentation with relative location prior.
\newblock International Journal of Computer Vision \textbf{80}(3) (2008)
  300--316

\bibitem{rohrbach2013translating}
Rohrbach, M., Qiu, W., Titov, I., Thater, S., Pinkal, M., Schiele, B.:
\newblock Translating video content to natural language descriptions.
\newblock In: Computer Vision (ICCV), 2013 IEEE International Conference on,
  IEEE (2013)  433--440

\bibitem{yao2010modeling}
Yao, B., Fei-Fei, L.:
\newblock Modeling mutual context of object and human pose in human-object
  interaction activities.
\newblock In: Computer Vision and Pattern Recognition (CVPR), 2010 IEEE
  Conference on, IEEE (2010)  17--24

\bibitem{maji2011action}
Maji, S., Bourdev, L., Malik, J.:
\newblock Action recognition from a distributed representation of pose and
  appearance.
\newblock In: Computer Vision and Pattern Recognition (CVPR), 2011 IEEE
  Conference on, IEEE (2011)  3177--3184

\bibitem{gupta2009observing}
Gupta, A., Kembhavi, A., Davis, L.S.:
\newblock Observing human-object interactions: Using spatial and functional
  compatibility for recognition.
\newblock Pattern Analysis and Machine Intelligence, IEEE Transactions on
  \textbf{31}(10) (2009)  1775--1789

\bibitem{yao2010grouplet}
Yao, B., Fei-Fei, L.:
\newblock Grouplet: A structured image representation for recognizing human and
  object interactions.
\newblock In: Computer Vision and Pattern Recognition (CVPR), 2010 IEEE
  Conference on, IEEE (2010)  9--16

\bibitem{ramanathan2015learning}
Ramanathan, V., Li, C., Deng, J., Han, W., Li, Z., Gu, K., Song, Y., Bengio,
  S., Rossenberg, C., Fei-Fei, L.:
\newblock Learning semantic relationships for better action retrieval in
  images.
\newblock In: Proceedings of the IEEE Conference on Computer Vision and Pattern
  Recognition. (2015)  1100--1109

\bibitem{guadarrama2013youtube2text}
Guadarrama, S., Krishnamoorthy, N., Malkarnenkar, G., Venugopalan, S., Mooney,
  R., Darrell, T., Saenko, K.:
\newblock Youtube2text: Recognizing and describing arbitrary activities using
  semantic hierarchies and zero-shot recognition.
\newblock In: Computer Vision (ICCV), 2013 IEEE International Conference on,
  IEEE (2013)  2712--2719

\bibitem{regneri2013grounding}
Regneri, M., Rohrbach, M., Wetzel, D., Thater, S., Schiele, B., Pinkal, M.:
\newblock Grounding action descriptions in videos.
\newblock Transactions of the Association for Computational Linguistics
  \textbf{1} (2013)  25--36

\bibitem{thomason2014integrating}
Thomason, J., Venugopalan, S., Guadarrama, S., Saenko, K., Mooney, R.:
\newblock Integrating language and vision to generate natural language
  descriptions of videos in the wild.
\newblock In: Proceedings of the 25th International Conference on Computational
  Linguistics (COLING), August. (2014)

\bibitem{yao2012describing}
Yao, J., Fidler, S., Urtasun, R.:
\newblock Describing the scene as a whole: Joint object detection, scene
  classification and semantic segmentation.
\newblock In: Computer Vision and Pattern Recognition (CVPR), 2012 IEEE
  Conference on, IEEE (2012)  702--709

\bibitem{kulkarni2011baby}
Kulkarni, G., Premraj, V., Dhar, S., Li, S., Choi, Y., Berg, A.C., Berg, T.L.:
\newblock Baby talk: Understanding and generating image descriptions.
\newblock In: Proceedings of the 24th CVPR, Citeseer (2011)

\bibitem{zitnick2013learning}
Zitnick, C.L., Parikh, D., Vanderwende, L.:
\newblock Learning the visual interpretation of sentences.
\newblock In: Computer Vision (ICCV), 2013 IEEE International Conference on,
  IEEE (2013)  1681--1688

\bibitem{gupta2008beyond}
Gupta, A., Davis, L.S.:
\newblock Beyond nouns: Exploiting prepositions and comparative adjectives for
  learning visual classifiers.
\newblock In: Computer Vision--ECCV 2008.
\newblock Springer (2008)  16--29

\bibitem{kumar2010efficiently}
Kumar, M.P., Koller, D.:
\newblock Efficiently selecting regions for scene understanding.
\newblock In: Computer Vision and Pattern Recognition (CVPR), 2010 IEEE
  Conference on, IEEE (2010)  3217--3224

\bibitem{russell2006using}
Russell, B.C., Freeman, W.T., Efros, A., Sivic, J., Zisserman, A.,  et~al.:
\newblock Using multiple segmentations to discover objects and their extent in
  image collections.
\newblock In: Computer Vision and Pattern Recognition, 2006 IEEE Computer
  Society Conference on. Volume~2., IEEE (2006)  1605--1614

\bibitem{farhadi2010every}
Farhadi, A., Hejrati, M., Sadeghi, M.A., Young, P., Rashtchian, C.,
  Hockenmaier, J., Forsyth, D.:
\newblock Every picture tells a story: Generating sentences from images.
\newblock In: Computer Vision--ECCV 2010.
\newblock Springer (2010)  15--29

\bibitem{berg2012understanding}
Berg, A.C., Berg, T.L., Daume~III, H., Dodge, J., Goyal, A., Han, X., Mensch,
  A., Mitchell, M., Sood, A., Stratos, K.,  et~al.:
\newblock Understanding and predicting importance in images.
\newblock In: Computer Vision and Pattern Recognition (CVPR), 2012 IEEE
  Conference on, IEEE (2012)  3562--3569

\bibitem{hoiem2008putting}
Hoiem, D., Efros, A.A., Hebert, M.:
\newblock Putting objects in perspective.
\newblock International Journal of Computer Vision \textbf{80}(1) (2008)  3--15

\bibitem{fang2014captions}
Fang, H., Gupta, S., Iandola, F., Srivastava, R., Deng, L., Doll{\'a}r, P.,
  Gao, J., He, X., Mitchell, M., Platt, J.,  et~al.:
\newblock From captions to visual concepts and back.
\newblock arXiv preprint arXiv:1411.4952 (2014)

\bibitem{chang2014semantic}
Chang, A.X., Savva, M., Manning, C.D.:
\newblock Semantic parsing for text to 3d scene generation.
\newblock ACL 2014 (2014) ~17

\bibitem{schustergenerating}
Schuster, S., Krishna, R., Chang, A., Fei-Fei, L., Manning, C.D.:
\newblock Generating semantically precise scene graphs from textual
  descriptions for improved image retrieval.
\newblock In: Proceedings of the Fourth Workshop on Vision and Language (VL15).
  (2015)

\bibitem{choi2013understanding}
Choi, W., Chao, Y.W., Pantofaru, C., Savarese, S.:
\newblock Understanding indoor scenes using 3d geometric phrases.
\newblock In: Computer Vision and Pattern Recognition (CVPR), 2013 IEEE
  Conference on, IEEE (2013)  33--40

\bibitem{krishnavisualgenome}
Krishna, R., Zhu, Y., Groth, O., Johnson, J., Hata, K., Kravitz, J., Chen, S.,
  Kalantidis, Y., Li, L.J., Shamma, D.A., Bernstein, M., Fei-Fei, L.:
\newblock Visual genome: Connecting language and vision using crowdsourced
  dense image annotations.
\newblock In: International Journal of Computer Vision. (2016)

\bibitem{girshick14CVPR}
Girshick, R., Donahue, J., Darrell, T., Malik, J.:
\newblock Rich feature hierarchies for accurate object detection and semantic
  segmentation.
\newblock In: Computer Vision and Pattern Recognition. (2014)

\bibitem{simonyan2014very}
Simonyan, K., Zisserman, A.:
\newblock Very deep convolutional networks for large-scale image recognition.
\newblock arXiv preprint arXiv:1409.1556 (2014)

\bibitem{alexe2012measuring}
Alexe, B., Deselaers, T., Ferrari, V.:
\newblock Measuring the objectness of image windows.
\newblock Pattern Analysis and Machine Intelligence, IEEE Transactions on
  \textbf{34}(11) (2012)  2189--2202

\bibitem{oliva2001modeling}
Oliva, A., Torralba, A.:
\newblock Modeling the shape of the scene: A holistic representation of the
  spatial envelope.
\newblock International journal of computer vision \textbf{42}(3) (2001)
  145--175

\bibitem{lowe2004distinctive}
Lowe, D.G.:
\newblock Distinctive image features from scale-invariant keypoints.
\newblock International journal of computer vision \textbf{60}(2) (2004)
  91--110

\end{thebibliography}
\end{document}